\documentclass[10pt,twocolumn,letterpaper]{article}

\usepackage[pagenumbers]{cvpr} 

\definecolor{cvprblue}{rgb}{0.21,0.49,0.74}
\usepackage[pagebackref,breaklinks,colorlinks,allcolors=cvprblue]{hyperref}
\usepackage{graphicx}	
\usepackage{hyperref}
\usepackage{url}
\usepackage{multirow}
\usepackage{makecell}
\usepackage{tabularray}
\usepackage{enumitem}

\def\confName{CVPR}
\def\confYear{2026}

\title{MAViD: A Multimodal Framework for Audio-Visual Dialogue Understanding and Generation}

\author{Youxin Pang\textsuperscript{1,2}\footnotemark[1] \footnotemark[2] \quad Jiajun Liu\textsuperscript{2}\footnotemark[1] \quad Lingfeng Tan\textsuperscript{2}\footnotemark[1] \quad Yong Zhang\textsuperscript{2}\footnotemark[3] \quad Feng Gao\textsuperscript{2} \\ Xiang Deng\textsuperscript{1,2}  \quad  Zhuoliang Kang\textsuperscript{2} \quad Xiaoming Wei\textsuperscript{2} \quad Yebin Liu\textsuperscript{1}\footnotemark[3] \\ \textsuperscript{1}Tsinghua University\\
\textsuperscript{2}Meituan\\ \\
}

\begin{document}

\twocolumn[{
\maketitle
\begin{center}
    \captionsetup{type=figure}
    \includegraphics[width=1.\linewidth]{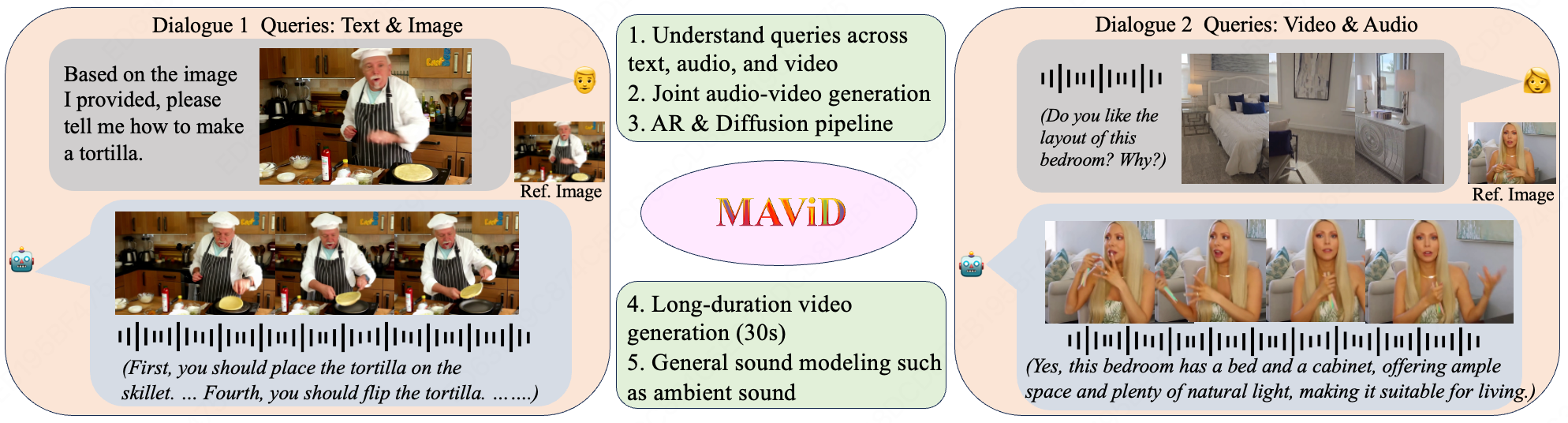}
    \caption{We present MAViD, a novel dialogue framework capable of understanding multimodal interactions across text, audio, and video, and generating highly realistic, human-like, long-duration synchronized audio–visual content. 
    Specifically, MAViD is capable of producing approximately 30-second videos in a single inference, whereas other DiT-based methods~\cite{low2025ovi,wang2025universe} can only generate 5-second clips.
    The figure illustrates two dialogue examples capable of handling different combinations of input modalities, where Ref.Image represents an optional reference image. 
    In dialogue 1, the model understands the input image and allows the person in the image to speak while performing corresponding actions.
    In dialogue 2, a human image can be designated as the response agent.
}
    \label{fig:teaser}
\end{center}
}]

\renewcommand{\thefootnote}{\fnsymbol{footnote}}
\footnotetext[1]{Equal contribution}
\footnotetext[2]{Work done during the internship at Meituan}
\footnotetext[3]{Corresponding authors.}

\begin{abstract}
We propose MAViD, a novel \textbf{M}ultimodal framework for \textbf{A}udio-\textbf{Vi}sual \textbf{D}ialogue understanding and generation.
Existing approaches primarily focus on non-interactive systems and are limited to producing constrained and unnatural human speech.
The primary challenge of this task lies in effectively integrating understanding and generation capabilities, as well as achieving seamless multimodal audio-video fusion.
To solve these problems, we propose a Conductor–Creator architecture that divides the dialogue system into two primary components.
The Conductor is tasked with understanding, reasoning, and generating instructions by breaking them down into motion and speech components, thereby enabling fine-grained control over interactions.
The Creator then delivers interactive responses based on these instructions.
Furthermore, to address the difficulty of generating long videos with consistent identity, timbre, and tone using dual DiT structures, the Creator adopts a structure that combines autoregressive (AR) and diffusion models. 
The AR model is responsible for audio generation, while the diffusion model ensures high-quality video generation.
Additionally, we propose a novel fusion module to enhance connections between contextually consecutive clips and modalities, enabling synchronized long-duration audio-visual content generation.
Extensive experiments demonstrate that our framework can generate vivid and contextually coherent long-duration dialogue interactions and accurately interpret users' multimodal queries.
\end{abstract}
\section{Introduction}\label{sec:intro}
Digital human dialogue interaction is a foundational aspect in computer vision, where enabling seamless multimodal communication between humans and digital agents is vital for downstream applications such as virtual assistants.
The core of interaction lies in multimodal understanding and generation capabilities.
Specifically, it involves the ability to understand users' inquiries across text, video, and audio modalities, and to provide appropriate responses.
Early on, such multimodal interaction is primarily expressed through text output~\cite{li2022blip,alayrac2022flamingo}, which limits their interaction scenarios.
Recently, numerous methods~\cite{xu2025qwen2,coreteam2025mimoaudio,ding2025kimi,ai2025ming, tong2025interactiveomni,huang2025step,higgsaudio2025,xie2024mini} have expanded multimodal interaction by incorporating the audio modality.
These methods feature speech generation capabilities, thereby enhancing the overall interactive experience.



In addition to text and audio, the ability to simultaneously generate visuals is an essential aspect of multimodal interaction, which is one of our core objectives.
Theoretically, the two-stage way of appending an audio-to-video module~\cite{zhang2024audio,pham2025spa2v} could equip the aforementioned methods with this capability.
For example, X-Streamer~\cite{xie2025x} proposes a similar approach that uses hidden states of generated text and audio as conditions for video generation.
However, the two-stage methods face several significant challenges.
Firstly, current MLLM methods often produce monotonous and evenly pitched speech, lacking human-like expressiveness.
While human-like TTS can convert text into vivid speech, the two-stage process of generating audio followed by video faces challenges in handling authentic general sounds, such as sound effects and real environmental noise. 
Consequently, this leads to inadequate alignment of visuals with these sounds.
Therefore, we explore the pipeline for joint audio-video generation, instead of text-audio generation.

In this paper, we introduce a novel dialogue framework capable of understanding multimodal interactions across text, audio, and video, and generating highly realistic, human-like, long-duration synchronized audio–visual content.
Specifically, we design a Conductor–Creator architecture that decouples understanding and generation into two collaborative components.
The Conductor is responsible for providing global textual instructions, while the Creator unifies detailed audio–visual content generation.
For Conductor, existing methods~\cite{xu2025qwen2} mainly generate speech-oriented text~\cite{xie2025x}, which limits the realism and naturalness of the visual results.
Therefore, we further divide the instructions generated by our Conductor into speech-oriented and motion-oriented textural instructions, enhancing fine-grained control over dynamic details.
Specifically, the speech instructions deliver essential auditory cues, whereas the motion instructions provide visual cues from the context.
Together, these guide the creation of realistic and compelling audio–visual content.

Another challenge lies in jointly modeling audio-visual modalities within the Creator to generate interactive outputs.
Recently, numerous methods~\cite{wang2025animate,low2025ovi,liu2025javisdit,haji2025av,wang2025universe,wang2025audiogen,zhang2025uniavgen} involve joint audio-video generation, with the dual DiT structure being popularly used to process audio and video.
However, this structure can only generate one audio-visual clip at a time, which increases the complexity of long-duration video generation.
It is particularly challenging to maintain consistency in identity, timbre, and tone across consecutive clips.
Therefore, we explore an alternative joint network based on autoregressive (AR) and diffusion models. 
AR models inherently possess the ability to model long sequences and multimodal content, while diffusion is used to maintain high visual quality.
Most current joint networks~\cite{xie2025show,zhou2024transfusion} focus primarily on text and visual modalities, utilizing self-attention to model different modalities within sequences.
However, this approach is insufficient to integrate text-audio-video content, so we design specialized attention fusion modules to establish connections between contextually consecutive clips and modalities.
Finally, our method facilitates long-duration audio–visual dialogue generation, capable of producing approximately 30-second videos in a single inference, whereas other DiT-based methods~\cite{low2025ovi,wang2025universe} can only generate 5-second clips.
Additionally, our method can model general environmental sounds, such as background noise during speech.

With the proposed Conductor–Creator architecture, our interactive dialogue framework can understand and generate multimodal content across text, audio, and video.
Through the decoupling of instructions and the integration of modality fusion modules, our method generates highly realistic and human-like synchronized audio–visual content.
This advances exploration of multimodal technologies based on the AR and diffusion frameworks, laying a solid foundation for constructing intelligent digital human agents.
We summarize our contributions as follows,
\begin{itemize}
    \item A novel dialogue framework that can understand multimodal interactions across text, audio, and video, and generate highly human-like, long-duration synchronized audio–visual content, including general sounds such as environmental noise.
    \item A novel Conductor module that provides global textual instructions, which are further divided into speech instructions and motion instructions, thereby enabling fine-grained control and enhancing realism.
    \item A novel Creator for joint audio–video generation that integrates the advantages of AR and diffusion models.
    It employs a carefully designed fusion module to integrate contextually consecutive clips and modalities, enabling the generation of long-duration content with more consistent identity, timbre, and tone.
\end{itemize}

\section{Related Work}\label{sec:related}

\noindent{\textbf{Multimodal Interaction.}} 
Recently, an increasing number of methods~\cite{xu2025qwen2,coreteam2025mimoaudio,ding2025kimi,ai2025ming, tong2025interactiveomni,huang2025step,higgsaudio2025,xie2024mini,fu2025vita,geng2025osum} are embedding audio into multimodal interactions, highlighting the capabilities of speech input and output.
For example, MiMo-Audio~\cite{coreteam2025mimoaudio} performs well across open-source benchmarks in audio comprehension, spoken dialogue, and instruct-TTS tasks.
Overall, these methods typically consist of an audio tokenizer, an audio LLM, and an audio detokenizer.
The pipeline begins with the audio tokenizer, which converts raw audio signals into discrete or continuous tokens. 
These tokens are then jointly modeled by the audio LLM with other modalities. 
Finally, the detokenizer translates audio tokens back into actual audio signals.
However, these methods can only generate audio and text, lacking the capability to generate visual signals.

\noindent{\textbf{Audio to Video Generation.}}
Recently, some methods~\cite{zhang2024audio,pham2025spa2v,wang2025keyvid,zhang2025scaling,yariv2024diverse} have explored using audio as a driving signal for video generation.
Specifically, AVSyncD~\cite{zhang2024audio} animates static images with audio-guided motions and introduces a dataset of synchronized audio–visual events. 
SpA2V~\cite{pham2025spa2v} approaches the task in two stages: first generating a layout using audio, then guiding video generation, thereby simplifying the task and enhancing video generation controllability. 
In addition, numerous methods~\cite{yang2025infinitetalk,pang2023dpe,zhao2025x,deng2025stereo} focus on generating talking heads.
Specifically, they seek to produce lip-synced videos from a single portrait image, driven by audio signals.
Integrating these techniques with audio-based MLLMs enables multimodal audio–visual interaction, yet this two-stage approach encounters several challenges.
Firstly, current MLLM methods often produce monotonous and evenly pitched speech, devoid of human-like expressiveness.
While human-like TTS can convert text into vivid speech, the two-stage process of generating audio followed by video faces challenges in handling authentic general sounds, such as sound effects and real environmental noise. 
Consequently, this leads to inadequate alignment of visuals with these sounds.

\noindent{\textbf{Joint Audio-Video Generation.}}
Recently, various methods~\cite{wang2025animate,low2025ovi,liu2025javisdit,haji2025av,wang2025universe,wang2025audiogen,hayakawa2024mmdisco,zhao2025uniform,liu2024syncflow,xing2024seeing,wang2025audiogen} have been developed to generate audio and video jointly from text inputs, which advances multimodal generation capabilities.
For example, OVI~\cite{low2025ovi} utilizes a dual DiT structure for separate audio and video generation, aligning the modalities with a fusion module.
Overall, these methods can model general sounds and achieve high-quality audio-visual synchronization.
However, they lack multimodal understanding and serve solely as generators.
Furthermore, they can only generate one audio-visual clip at a time, increasing the complexity of long-duration video generation. 
This makes it particularly challenging to ensure consistency in identity, timbre, and tone across consecutive clips.

\noindent{\textbf{AR and Diffusion Model.}} 
From the framework perspective, most visual and audio generation methods~\cite{wan2025wan,liu2024audioldm,gao2025seedance,shin2025deeply,team2025longcat} adopt diffusion-based frameworks, such as DiT~\cite{peebles2023scalable}, to produce high-quality results. 
In contrast, text generation~\cite{brown2020language,yang2019xlnet,touvron2023llama} frequently employs autoregressive (AR) frameworks, which are advantageous for modeling long sequences. 
Recently, an increasing number of methods~\cite{wang2024emu3,wu2025janus,chen2025janus} have leveraged AR frameworks to achieve unified modeling of text and visual modalities. 
By representing text and visuals as a single token sequence, joint generation can be accomplished using a single transformer. 
However, due to the complexity of the visual modality, simple discrete token representations lead to suboptimal visual generation quality~\cite{agarwal2025cosmos}. 
Consequently, many approaches~\cite{xie2025show,zhou2024transfusion,xie2024show} are integrating AR with diffusion to combine their strengths for unified multimodal modeling and high-quality visual generation.
However, they utilize self-attention and bidirectional attention across visual regions to model text-visual modalities, which is insufficient for integrating text-audio-video modalities.
\section{Method}\label{sec:met}
\begin{figure*}[tp]
    \centering
    \includegraphics[width=1.\linewidth]{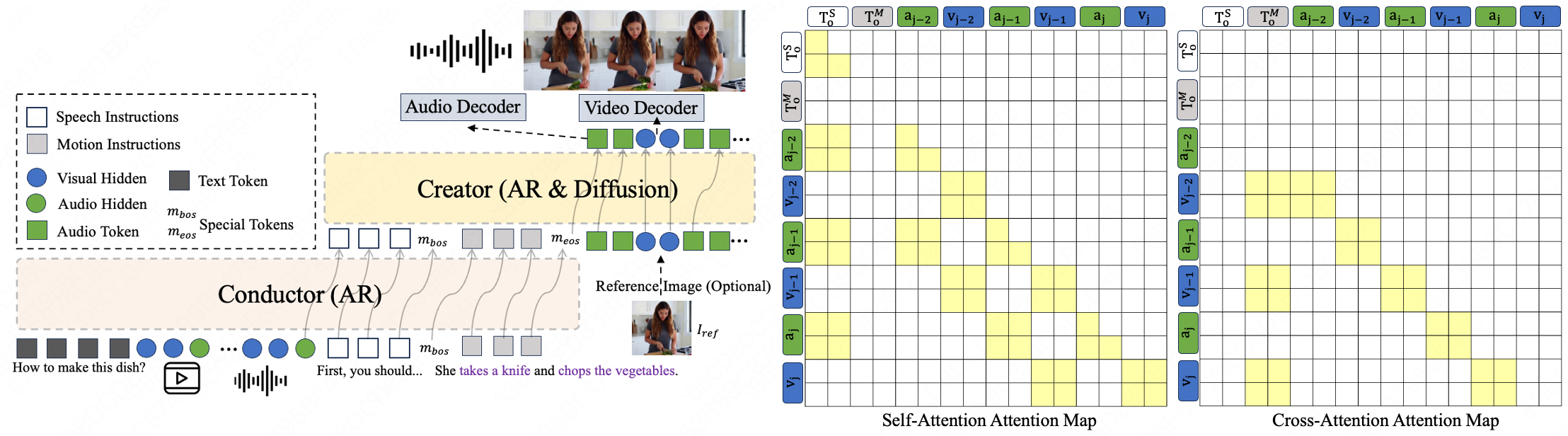}
    \caption{Overview of MAViD. Left: The Conductor-Creator architecture. The Conductor takes users' inquiries across text, audio, and video as input, understands them, and outputs textual instructions.
    To achieve fine-grained control over video generation, these textual instructions are further decoupled into speech-oriented and motion-oriented instructions. 
    In Creator, the decoupled instructions guide the joint audio-video generation. Specifically, we employ a structure combining autoregressive (AR) and diffusion models to model long sequences and maintain visual quality.
    Right: To ensure consistency and coherence in long-sequence joint audio-video generation, we propose a fusion module that integrates features from both AR and diffusion. The figure illustrates the use of attention for interaction among interleaved audio and video clips, with yellow parts indicating where attention needs to be computed.
}
    \label{fig:pipeline}
\end{figure*}

The overall pipeline of MAViD is illustrated in Fig.~\ref{fig:pipeline}. 
Our goal is to construct a multimodal audio-visual dialogue interaction framework that outputs appropriate text-audio-video pairs $O = \{T_{o}, A_{o}, V_{o}\}$ based on users' inquiries, which include text $T_{in}$, audio $A_{in}$, and video $V_{in}$ (where the image is treated as a single-frame video for consistency). 
Specifically, the entire framework is organized around the Conductor–Creator architecture.
The Conductor is responsible for providing global textual instructions, while the Creator unifies detailed audio–visual content generation.

We first introduce the Conductor in Sec.~\ref{sec:Conductor}. 
For fine-grained control over motion and speech, the Conductor’s instructions are further divided into speech and motion instructions.
The speech instructions deliver essential auditory cues, whereas the motion instructions deliver visual cues from the environment.
Together, these guide the creation of realistic and compelling audio–visual content.
Then, we detail the Creator in Sec.~\ref{sec:Creator}, which is a structure for combining AR with diffusion. 
Moreover, to integrate three distinct modalities within a single transformer, we carefully design an attention fusion module for multimodal modeling and interaction.
After that, we discuss the training and inference strategies in Sec.~\ref{sec:train}.

\subsection{Conductor}
\label{sec:Conductor}
In dialogue interaction frameworks, understanding the users' multimodal inquiries, which consist of text, audio, and video, is crucial for providing appropriate feedback and generating corresponding responses.
As shown in Fig.~\ref{fig:pipeline}, we adopt the Thinker part from Qwen2.5-omni~\cite{xu2025qwen2} as our baseline. 
Our Conductor consists of three encoders $(E_{T}^{c}, E_{A}^{c}, E_{V}^{c})$ corresponding to $\{T_{in}, A_{in}, V_{in}\}$ and a transformer decoder.
Encoders encode all three modalities into corresponding hidden embeddings or tokens, subsequently, transformers use these features as inputs, understand the semantic information, and reply to the textual information $T_{o}$ through next token prediction.
The entire process can be modeled as:
\begin{equation}
    T_{o} = M_{c}(E_{T}^{c}(T_{in}), E_{A}^{c}(A_{in}), E_{V}^{c}(V_{in})),
    \label{eq:Conductor1}
\end{equation}
where $M_{c}$ means the transformer decoder in Conductor.
Specifically, we employ the same three encoders as those used in Qwen2.5-omni Thinker.

Similarly, Qwen2.5-omni~\cite{xu2025qwen2} and X-Streamer~\cite{xie2025x} propose a Thinker module to handle these inquiries and generate textual responses. 
InteractiveOmni~\cite{tong2025interactiveomni} unifies audio and text responses within a single model, thereby achieving their joint generation.
However, the generated textual responses do not account for motion information, which is critical for enhancing the human-likeness and realism of interactions.
For example, when replying with a message of agreement, one should be able to simultaneously perform a nodding action. 
This requires the model to not only perform speech responses, but also carry out semantic and contextual analysis, as well as provide more detailed interactive responses.



Therefore, we propose to decouple the instructions generated by Conductor into motion and speech directives.
The speech instructions $T_{o}^{S}$ deliver essential auditory cues, whereas the motion instructions $T_{o}^{M}$ deliver visual cues from the environment and context.
Based on this, Eq.~\ref{eq:Conductor1} becomes:
\begin{equation}
    (T_{o}^{S}, T_{o}^{M}) = M_{c}(E_{T}^{c}(T_{in}), E_{A}^{c}(A_{in}), E_{V}^{c}(V_{in})).
\end{equation}
To achieve this, we reconstruct the input sequence.
As shown in Fig.~\ref{fig:pipeline}, in the condition part, we place the text at the beginning, followed by interleaved audio and video sequences to strengthen the connection between them.
In the generated part, we use special tokens to separate motion cues from speech cues:
\begin{equation}
    [T_{o1}^{s} \ \ \ \ T_{o2}^{s} \ \ \ \ ... \ \ \ \ m_{bos} \ \ \ \ T_{o1}^{m} \ \ \ \ T_{o2}^{m} \ \ \ \ ... \ \ \ \ m_{eos}],
    \label{eq:mmm}
\end{equation}
where $T_{oi}^{s}$ and $T_{oi}^{m}$ respectively represent the $i$-th text token in speech directives and motion directives, $m_{bos}$ and $m_{eos}$ are special tokens to represent the start and end of the motion directives, respectively.



Through the aforementioned modeling approach, our Conductor can generate instructions across multiple levels after understanding multimodal inquiries, achieving fine-grained control over motion and speech.
Additionally, to retain strong understanding capabilities while decoupling instructions, we employ a mixed training strategy in Sec.~\ref{sec:train}.
Finally, compared to existing methods~\cite{xu2025qwen2,xie2025x,tong2025interactiveomni}, the Conductor produces instructions that are more detailed, comprehensive, and enriched with dynamic information.

\subsection{Creator}
\label{sec:Creator}
The goal of Creator is to transform the feedback from Conductor into interaction information that users can recognize. 
Methods such as Qwen2.5-omni~\cite{xu2025qwen2}, InteractiveOmni~\cite{tong2025interactiveomni}, and MiMo-Audio~\cite{coreteam2025mimoaudio} ultimately present their outputs in the form of text and speech.
In addition to text and audio, the ability to simultaneously generate visuals is an essential aspect of multimodal interaction.
A straightforward approach is to use a two-stage method for generating audio and video. 
However, this method fails to produce vivid human speech or model general environmental sounds.


\begin{table*}[ht]
\centering
\caption{Performance evaluation for the Conductor on multimodal understanding tasks (Image $\to$ Text). To ensure a fair comparison, we test all methods using VLMEvalKit~\cite{duan2024vlmevalkit} under the same settings. From the metrics, it is evident that our Conductor effectively retains the understanding capabilities of the baseline, achieving results comparable to other methods.}
\label{tab:img2text}
\begin{tabular}{l|ccccccc}
\toprule
\textbf{Model} & MMstar & MMMU & MMBench-V1.1 & MME & MUIRBench & HallusionBench & RealWorldQA\\
\midrule
Qwen2.5-Omni-7B & 61.33 & 56.38 & \textbf{82.08} & \textbf{2249} & \textbf{59.04} & 46.27 & \textbf{70.06} \\
VITA-1.5 & 60.13 & 50.00 & 78.93 & 2201 & 55.69 & 44.90 & 66.67 \\
Ours (Conductor)  & \textbf{62.00} & \textbf{57.24} & 80.42 & 2244 & 56.50 & \textbf{46.90} & 66.93 \\
\bottomrule
\end{tabular}
\end{table*}

To address this, based on the output of Conductor, our Creator is designed to jointly generate audio synchronized videos:
\begin{equation}
    (A_{o}, V_{o}) = \mathcal{G}(T_{o}^{S}, T_{o}^{M}, I_{ref}),
\end{equation}
where $\mathcal{G}$ represents the entire Creator and $I_{ref}$ is optionally used as the reference image.
To achieve joint audio-video generation, one straightforward approach is to build the model based on DiT.
Recently, OVI~\cite{low2025ovi} utilizes twin DiT networks to independently process audio and video, using fusion modules for alignment.
Although the dual DiT structure effectively integrates two modalities, it can only generate one audio-visual clip at a time, increasing the complexity of long-duration video generation and making it difficult to ensure consistency in identity, timbre, and tone across consecutive clips.
Therefore, we explore an alternative approach based on autoregressive (AR) models. 
Firstly, AR models inherently possess the ability to model long sequences. 
Additionally, AR is more suitable for multimodal modeling~\cite{tan2025omni,wei2025univideo}, providing a foundation for integrating additional modalities such as human pose and camera in the future.

As depicted in Fig.~\ref{fig:pipeline}, we adopt the Talker framework in Qwen2.5-omni~\cite{xu2025qwen2} as our baseline.
The complete Creator comprises three encoders $(E_{T}^{g}, E_{A}^{g}, E_{V}^{g})$ dedicated to text, audio, and video modalities, along with a transformer decoder and two audio-video decoders.
Specifically, to encode general sound, $E_{A}^{g}$ employs the tokenizer in higgs-audio~\cite{higgsaudio2025}, whereas $E_{V}^{g}$ uses the VAE in Wan~\cite{wan2025wan}.
Considering that the visual modality is more complex than audio, we integrate a diffusion module into the AR baseline to handle visual sequences and maintain high visual quality.
Specifically, we embed the DiT block from Wan~\cite{wan2025wan} into our AR baseline.
For sequence modeling, we place the text at the beginning, followed by interleaved audio and video clips to strengthen the connection between them.
As shown in Fig.~\ref{fig:pipeline}, the entire framework operates in an autoregressive manner.
When generating the current clip, historical clips are used as conditions, thus laying the foundation for long-duration generation:
\begin{equation}
    AV_{i} = \mathcal{G}(AV_{i-1}, T_{o}^{S}, T_{o}^{M}, I_{ref}),
    \label{eq:his}
\end{equation}
where $AV_{i}$ denotes the $i$-th clip in audio and video clips.
It is worth noting that specifically within the audio clip, the next token prediction method is employed, while within the video clip, the diffusion pattern is utilized for overall denoising.

\textbf{Fusion module for Text, Video and Audio.}
To ensure consistency and coherence in long-sequence joint
audio-video generation, we design specialized fusion attention strategies to process text-audio-video content.
In each decoder layer, self-attention ($\text{SA}$) is primarily used to establish connections between different clips within the same modality, followed by cross-attention ($\text{CA}$) to create associations between different modalities.
As shown on the right side of Fig.~\ref{fig:pipeline}, when predicting $j$-th ($j \textgreater 1$) audio latent clip $a_{j}$, speech directives, historical audio clip $a_{j-1}$, and clean video clips $v_{j-1}$ are used as conditions:
\begin{equation}
    \hat{a}_{j} = \text{CA}(\text{SA}([T_{o}^{S} \circ a_{j-1} \circ a_{j}]), \mathcal{F}_{v}(\text{SA}(v_{j-1}))),
\end{equation}
where $\hat{a}_{j}$ is the hidden states after fusion, $x$ is the query and $y$ is the key/value in $\text{CA}(x, y)$, $[x \circ y]$ concatenates $x$ and $y$ along the sequence length dimension.
Notably, we find that injecting the entire $v_{j-1}$ using CA leads to a decline in performance, possibly due to the introduction of irrelevant early video information.
Therefore, $\mathcal{F}_{v}$ represents the injection of only the last 10 latents of $v_{j-1}$, which corresponds to approximately 40 frames.
When denoising the $j$-th ($j \textgreater 1$) noisy video latent clip $v_{j}$, we adopt the attention strategy as follows:
\begin{align}
   \hat{v}_{j} = 
            \text{CA}(\text{CA}(\text{SA}([v_{j-1} \circ v_{j}]), T_{o}^{M}), \mathcal{F}_{a}({a_{j}})),
\end{align}
where $\hat{v}_{j}$ is the hidden states after fusion.
To enhance the temporal consistency of audio-video, we use $\mathcal{F}_{a}$ to identify more relevant audio conditions.
Specifically, for each video latent, we first identify the corresponding temporal audio position from $a_{j}$ and then embed the first four audio tokens, corresponding to approximately 40ms of audio.
Through the collaborative efforts of AR, diffusion and fusion module, our Creator can generate long-duration synchronized audio-video content while modeling general sounds such as background noise.

\begin{table*}[t]
    \centering
    \small
    \caption{Quantitative comparison of our Creator with different methods. JavisDiT lacks the capability to process human speech and is therefore not included in the calculation of lip synchronization metrics (LS). 
    Notably, although our method can generate longer videos in a single inference, we adopt the 5-second generation setting used by other methods to ensure fairness.}
    \label{tab:method_comparison}
    \resizebox{1\textwidth}{!}{
    \begin{tabular}{l c c c c c c c c c c c}
        \toprule
        \multirow{2}{*}{\textbf{Methods}} & \multirowthead{2}[2.5pt]{\textbf{\thead{Parameters \\ (S+V)}}} & \multicolumn{4}{c}{\textbf{Audio Quality}} & \multicolumn{3}{c}{\textbf{Video Quality}} & \multicolumn{3}{c}{\textbf{Audio-Video Consistency}} \\
        \cmidrule(lr){3-6} \cmidrule(lr){7-9} \cmidrule(lr){10-12}
        & & PQ($\uparrow$) & CU($\uparrow$) & PC($\uparrow$) & WER($\downarrow$) & SC($\uparrow$) & DD($\uparrow$) & IQ($\uparrow$) & LS($\uparrow$) & TC($\uparrow$) & SAC($\uparrow$) \\
        \midrule
        \multicolumn{10}{l}{\textit{\footnotesize \textbf{Two-stage Generation}}} \\
        Wan-S2V~\cite{gao2025wan} & 16.6B & \textbf{7.698} & \textbf{7.178} & 1.432 & \textbf{0.102} & \textbf{0.978} & 0.559 & 0.699 &  \textbf{7.400} &  0.612 & 0.393 \\
        \midrule
        \multicolumn{10}{l}{\textit{\footnotesize \textbf{Joint Generation} }} \\
        JavisDiT~\cite{liu2025javisdit} & 3.7B & 4.945 & 3.868 & \underline{2.138} & 0.907 & 0.963 & 0.214 & 0.661 & - & 0.587 & 0.471 \\
        Universe-1~\cite{wang2025universe} & 7.1B & 4.116 & 3.789 & 1.802 & 0.268 & 0.968 & 0.524 & 0.696 & 1.934 & 0.526 & 0.302 \\
        Ovi~\cite{low2025ovi} & 10.9B & 5.713 & \underline{5.880} & 1.728 & \underline{0.121} & 0.966 & \underline{0.667} & \textbf{0.705} &  \underline{7.113} & \underline{0.744} & \underline{0.570} \\
         Ours & 8.0B & \underline{6.007} & 5.609 & \textbf{2.255} & 0.126 & \underline{0.969} & \textbf{0.828} & \underline{0.702} & 6.551 & \textbf{0.767} & \textbf{0.595} \\
        \bottomrule
    \end{tabular}
    }
\end{table*}
\subsection{Training and Inference Strategy}
\label{sec:train}
\textbf{Training.}
MAViD is trained with a three-stage approach. 
In the first stage, we train the Conductor in an end-to-end way with cross-entropy loss:
\begin{equation}
    \mathcal{L}_{AR} = - \sum_{i=1}^{L} \log p (q_{i} | q_{<i}, u),
    \label{eq:loss_en}
\end{equation}
where $L$ is the target sequence length, and $q_{i}$ is the ${i}$-th token in the sequence with $u$ denoting the conditions.
Specifically, we use the Thinker in Qwen2.5-omni~\cite{xu2025qwen2} as baseline for fine-tuning.
To decouple outputs while retaining the strong understanding capabilities of the baseline, we construct a diverse dataset. 
First, for datasets not involving motions, such as simple audio QA used in Qwen2.5-omni, we place the response content into speech instructions and set motion instructions to null, preventing excessive decoupling that could compromise the baseline's capabilities.
Second, we collect a series of human-centered datasets including dialogue, Video QA, speech instruction, audio–video instruction, and audio–video QA, which are modeled according to Eq.~\ref{eq:mmm}.
Finally, by mixing these two types of data, the Conductor can preserve a more complete understanding capability while achieving target decoupling.
It is important to note that during inference, users may not necessarily input all three modalities simultaneously. 
Therefore, when constructing the training set, we incorporate various modality combinations to enhance generalization.
Please refer to our supplementary paper for more results.

The Creator is trained in the remaining two stages.
Specifically, in the second stage, we only train the AR baseline for audio generation using the loss defined in Eq.~\ref{eq:loss_en}.
In the last stage, we add DiT blocks to the AR baseline and then train the entire model end-to-end:
\begin{equation}
    \mathcal{L}_{all} = \mathcal{L}_{AR} + \mathcal{L}_{DIFF} ,
    \label{eq:vqloss}
\end{equation}
where $\mathcal{L}_{DIFF}$ represents the diffusion velocity loss.

\textbf{Inference.}
During the inference phase, users can input queries composed of any modality.
Firstly, the Conductor unifies the response to both speech directives and motion directives. 
In Creator, besides the directives generated by Conductor, an optional reference image $I_{ref}$ can also be used as input. 
Therefore, if $I_{ref}$ is provided, we replace the first frame with $I_{ref}$ as guidance when generating the first video clip.
Finally, we use the tokenizer decoder in higgs-audio~\cite{higgsaudio2025} to convert the generated audio features into real sound.
In addition, the delay pattern~\cite{higgsaudio2025} is applied to enable simultaneous audio token generation across codebooks, as the audio tokenizer contains multiple codebooks.
Additionally, the video is generated via VAE decoder in Wan~\cite{wan2025wan}.

When generating long-duration videos, we inject historical clips as conditions into the current clip using the fusion module described in Sec.~\ref{sec:Creator}. 
Notably, because audio and video prompts may vary within long sequences, $T_{o}^{S}$ and $T_{o}^{M}$ in Eq.~\ref{eq:his} represent the current clip's prompt.


\section{Experiments}\label{sec:expe}

\begin{figure*}[tp]
    \centering
    \includegraphics[width=1.\linewidth]{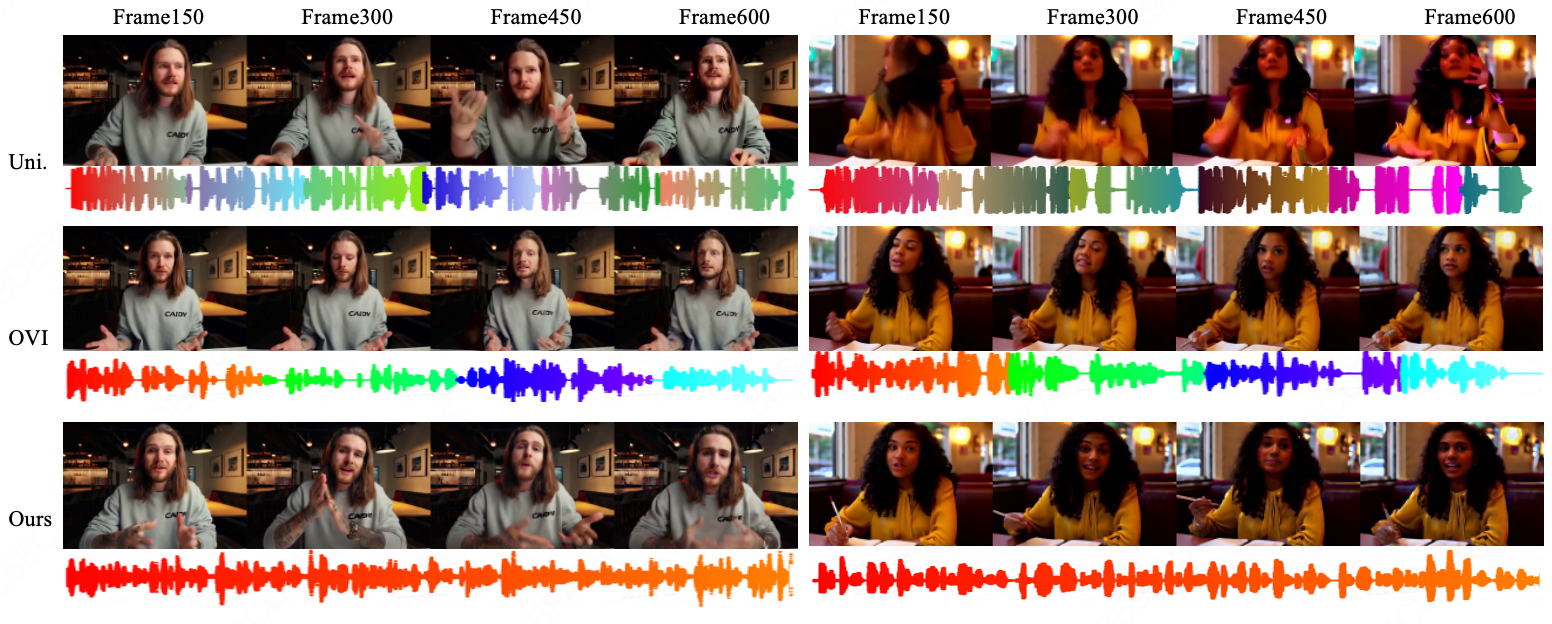}
    \caption{Visual comparison of long videos. For each method, we generate approximately 600 frames of video using the same audio and video prompts. Ovi~\cite{low2025ovi} and Universe-1~\cite{wang2025universe} (Uni.) employ multiple rounds of inference, using the last frame of the previous clip as the reference image. Different audio colors represent variations in timbre and tone.}
    \label{fig:comp_long}
\end{figure*}

\subsection{Settings}

\noindent{\textbf{Dataset.}}
Please refer to the supplementary materials for the dataset used in Conductor.
For the Creator, during the audio training phase, we collect over 10000 hours of audio data from the internet, primarily comprising human-centered audio, including speech, music, and some ambient sounds.
The data is annotated using Gemini-2.5-Pro, covering speaking content, gender, speaking tone, and description of ambient sounds.
During the audio-video joint training phase, we gather nearly 100w audio-video pairs, primarily focusing on movies, talking body videos, speeches, and human interaction scenes. 
The audio modality annotation is consistent with the audio training phase. 
For video annotation, Gemini-2.5-Pro is used to detail information such as human identity, background, scene, environment, and actions.

\noindent{\textbf{Benchmarks.}}
We conduct comprehensive evaluation of the proposed model, including independent comparisons of Conductor and Creator. 
Specifically, for Conductor, experiments are performed based on publicly available benchmarks to assess the multimodal understanding capabilities. 
Depending on the input modality, the evaluation is divided into Image $\to$ Text, Audio $\to$ Text, and Video $\to$ Text. 
For Image $\to$ Text, we employ seven benchmarks: MMstar~\cite{chen2024we}, MMMU~\cite{yue2024mmmu}, MMBench-V1.1~\cite{liu2024mmbench}, MME~\cite{fu2023mme}, MUIRBench~\cite{wang2024muirbench}, HallusionBench~\cite{guan2024hallusionbench}, and RealWorldQA~\cite{grok15}. 
Video $\to$ Text primarily includes Video-MME~\cite{fu2025video}, MLVU~\cite{zhou2025mlvu}, LongVideoBench~\cite{wu2024longvideobench}, and MVBench~\cite{li2024mvbench}. 
For Audio $\to$ Text, the evaluation includes a wide range of tasks across automatic speech recognition (ASR), audio understanding, audio reasoning, and voice-chatting.
Specifically, the ASR benchmarks include LibriSpeech~\cite{panayotov2015librispeech}, Fleurs~\cite{conneau2023fleurs}, and CommonVoice~\cite{ardila2019common}.
The audio understanding and reasoning includes MMAU~\cite{sakshi2024mmau}, VocalSound~\cite{gong2022vocalsound}, and MELD~\cite{poria2019meld}.
VoiceBench~\cite{chen2024voicebench} is used for voice-chatting.

For Creator, we conduct comparisons in two aspects: generation paradigm and network structure. 
(1) Two-stage generation paradigm: audio is generated first and then used to guide video generation. 
Specifically, we utilize CosyVoice 2~\cite{du2024cosyvoice} as TTS to convert speech text to audio, followed by integrating Wan-S2V~\cite{gao2025wan} for video generation. 
(2) Diffusion-based structures: A dual DiT structure is employed to model audio and video. 
We specifically compare our approach with JavisDiT~\cite{liu2025javisdit}, UniVerse-1~\cite{wang2025universe}, and OVI~\cite{low2025ovi}.
Following UniAVGen~\cite{zhang2025uniavgen}, we evaluate the metrics from three dimensions.
First, we adopt AudioBox-Aesthetics~\cite{tjandra2025meta}, which includes Production Quality (PQ), Production Complexity (PC), and Content Usefulness (CU) to evaluate audio quality.
Also, Whisper-largev3~\cite{radford2023robust} is used to compute the Word Error Rate (WER) of the generated audio.
Secondly, we use VBench~\cite{huang2024vbench} to evaluate video quality, including Subject Consistency (SC), Dynamic Degree (DD), and Imaging Quality (IQ).
Finally, we evaluate audio-video consistency, which includes lip synchronization (LS), timbre consistency (TC), and Scene-Audio Consistency (SAC).
For TC and SAV, we use Gemini-2.5-Pro.

\begin{table*}[t]
    \centering
    \small
    \caption{Quantitative comparison of our Creator with different methods on long video generation. We calculate the metrics for 18-second videos, which are generated by multiple inferences for other methods. }
    \label{tab:long}
    \resizebox{1\textwidth}{!}{
    \begin{tabular}{l c c c c c c c c c c c}
        \toprule
        \multirow{2}{*}{\textbf{Methods}} & \multirowthead{2}[2.5pt]{\textbf{\thead{Parameters \\ (S+V)}}} & \multicolumn{4}{c}{\textbf{Audio Quality}} & \multicolumn{3}{c}{\textbf{Video Quality}} & \multicolumn{3}{c}{\textbf{Audio-Video Consistency}} \\
        \cmidrule(lr){3-6} \cmidrule(lr){7-9} \cmidrule(lr){10-12}
        & & PQ($\uparrow$) & CU($\uparrow$) & PC($\uparrow$) & WER($\downarrow$) & SC($\uparrow$) & DD($\uparrow$) & IQ($\uparrow$) & LS($\uparrow$) & TC($\uparrow$) & SAC($\uparrow$) \\
        \midrule
        Universe-1~\cite{wang2025universe} & 7.1B & 3.763 & 3.641 & 1.863 & 0.271 & 0.910 & 0.521 & 0.673 & 1.895 & 0.511 & 0.305 \\
        Ovi~\cite{low2025ovi} & 10.9B & \underline{5.207} & \textbf{5.703} & 1.704 & \textbf{0.120} & \underline{0.951} & 0.665 & \textbf{0.692} &  \textbf{7.035} & \underline{0.712} & \underline{0.576} \\
         Ours (long) w/o fusion & 8.0B & 5.032 & 4.494 & \underline{2.046} & 0.158 & 0.944 & \underline{0.705} & \underline{0.688} & 3.738 & 0.699 & 0.568 \\
         Ours (long) & 8.0B & \textbf{5.687} & \underline{5.332} & \textbf{2.294} & \underline{0.129} & \textbf{0.954} & \textbf{0.754} & 0.679 & \underline{6.183} & \textbf{0.720} & \textbf{0.580} \\
        \bottomrule
    \end{tabular}
    }
\end{table*}
\subsection{Baseline Comparisons}
\textbf{The Conductor.}
Our Conductor is designed to understand multimodal queries and respond with textual instructions.
To enhance controllability and provide more human-like visual effects, we further decouple the generated instructions into motion instructions and speech instructions. 
The speech instructions are responsible for auditory interaction, while the motion instructions handle context interaction related to the environment. 

Overall, the objective is to decouple instructions while preserving the strong understanding capabilities of the baseline as much as possible.
Therefore, we conduct a comprehensive evaluation of the understanding capabilities.
Specifically, we compare our Conductor with methods Qwen2.5-omni~\cite{xu2025qwen2} and VITA-1.5~\cite{fu2025vita}.
Notably, to ensure fairness, all methods are tested under the same settings. 
Moreover, most evaluation benchmarks are simple multiple-choice questions or QA. 
Due to the mixed training strategy described in Sec.~\ref{sec:train}, the decoupling of instructions does not impact benchmark testing.
As shown in Tab.~\ref{tab:img2text},
our Conductor is comparable to the baseline in terms of image understanding capabilities, significantly retaining the strong understanding abilities of the baseline. 
Please refer to the supplementary materials for more details about video understanding and audio understanding.

\textbf{The Creator.}
Unlike the two-stage generation methods, our Creator performs joint audio-video generation in one stage.
Also, rather than using a dual DiT structure, our Creator employs a framework that combines an autoregressive model with diffusion.
The AR model is inherently suitable for multimodal and long-sequence modeling, while embedded DiT blocks ensure visual quality.
Therefore, our experiments compare the two types of methods mentioned above.

As shown in Tab.~\ref{tab:method_comparison}, the two-stage method achieves better audio results due to its use of dedicated and independent TTS for audio generation. 
However, in joint generation, our method and OVI~\cite{low2025ovi} achieve better audio results.
The sound generated by JavisDiT~\cite{liu2025javisdit} and Universe-1~\cite{wang2025universe} contains noticeable sharp noise, and JavisDiT lacks optimization for human speech.
From the perspective of visual quality, our generated results achieve the best subject consistency in joint generation. 
Additionally, based on the Dynamic Degree, our results exhibit stronger dynamic performance, whereas other methods tend to generate more static outcomes.
Consequently, the enhanced dynamics result in a slight decrease in image quality.
In terms of audio-video consistency, the dual DiT structure enables OVI to achieve better metrics by effectively aligning audio and visual features block-by-block at intermediate layers.
Universe-1 performs poorly due to the generation of abnormal noise. 
Nevertheless, from TC and SAV, our method demonstrates superior consistency in timbre and scene alignment.

\noindent{\textbf{Long Video.}}
Although the dual DiT structure effectively integrates audio and video content, it only generates one audio-video clip at a time. 
This increases the complexity of long-duration video generation, making it particularly challenging to ensure consistency in identity, timbre, and tone across consecutive clips.
Therefore, we explore an alternative joint network based on AR and diffusion models.
To enhance the coherence in long-duration generation, we propose a fusion module in Sec.~\ref{sec:Creator} to integrate intermediate features of AR and diffusion. 
As shown in Fig.~\ref{fig:comp_long}, we present the results of generating long-duration videos. 
Ovi and Universe-1 (Uni.) can only generate one clip per inference, so we perform four consecutive inferences, using the last frame of the previous clip as the reference image each time. 
Ultimately, all three methods generated approximately 600 frames. 
The color of the audio represents changes in timbre and tone. 
It can be observed that our method, due to generating all clips in one inference and the cumulative error of the AR model, results in gentle and slight changes.
For OVI, the lack of modeling historical clips leads to abrupt audio changes. 
Universe-1 generates a significant amount of sharp noise, reducing overall audio quality.
In Tab.~\ref{tab:long}, we calculate the metrics for three consecutive clips, and the conclusions are consistent with the visual results.

\begin{figure}[tp]
    \centering
    \includegraphics[width=1.\linewidth]{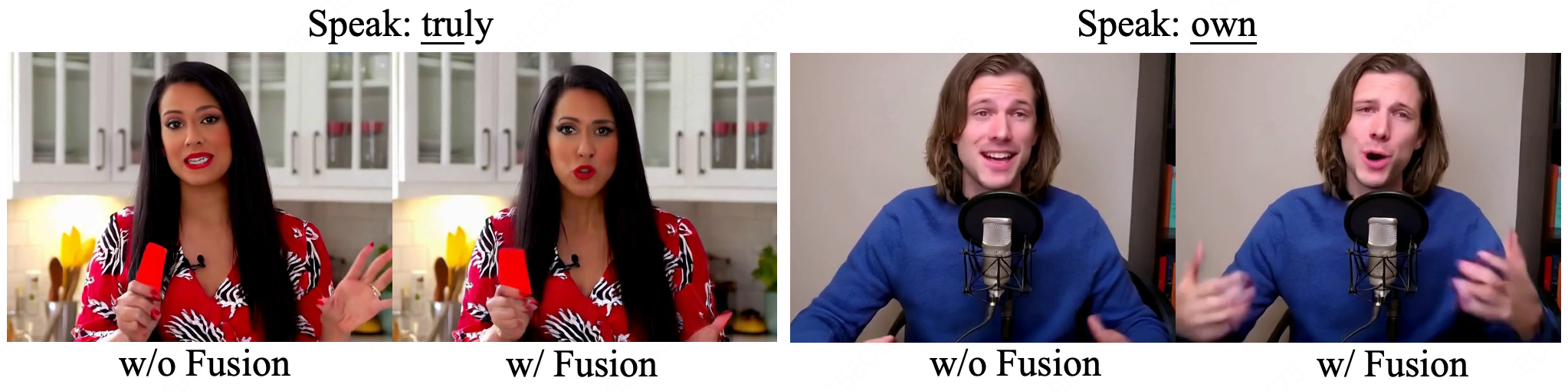}
    \caption{Ablation experiment of the fusion module.}
    \label{fig:abl}
    \vspace{-0.5 cm}
\end{figure}

\subsection{Ablation Study}
Recently, some multimodal methods~\cite{xie2024show,zhou2024transfusion} combine AR with diffusion models, using AR for text and diffusion for visual information. 
However, they utilize self-attention and bidirectional attentions across visual regions to model text-visual modalities.
We find that when dealing with text-audio-visual modalities, the aforementioned ways are insufficient. 
Therefore, we design a novel fusion module (Sec.~\ref{sec:Creator}) to enhance the connections between multimodal context clips, thereby supporting long-duration generation.
In the experiments, we compare with our Creator without using the fusion module (w/o fusion), generating 18-second videos in a single inference to evaluate the impact of context embedding.
As shown in Tab.~\ref{tab:long}, the capability of audio generation declines significantly when the fusion module is not utilized, resulting in lower audio-video consistency. 
Besides, as shown in Fig.~\ref{fig:abl}, the fusion module plays a crucial role in ensuring audio-video consistency.
\section{Conclusion}\label{sec:conclusion}

We introduced MAViD, a multimodal framework designed for audio-visual dialogue understanding and generation. 
By implementing the Conductor–Creator architecture, we effectively integrate understanding and generation capabilities, allowing for fine-grained control over interactions through the breakdown of instructions into motion and speech components.
The Creator employs a combination of AR and diffusion models to tackle the challenges of generating long videos with consistent identity, timbre, and tone. 
Additionally, the proposed novel fusion module facilitates synchronized long-duration audio-visual content generation by strengthening connections between contextually consecutive clips and modalities.
Finally, comprehensive quantitative and qualitative analyses demonstrate the effectiveness of our approach.
{
    \small
    \bibliographystyle{ieeenat_fullname}
    \bibliography{main}
}


\end{document}